%% file: main.tex
\documentclass[10pt,twocolumn,letterpaper]{article}

\usepackage{iccv}              

\input{preamble}

\definecolor{iccvblue}{rgb}{0.21,0.49,0.74}
\usepackage[pagebackref,breaklinks,colorlinks,allcolors=iccvblue]{hyperref}

\def\paperID{455} 
\def\confName{ICCV}
\def\confYear{2025}

\input{math_commands}

\usepackage{overpic}

\usepackage{multirow}
\usepackage{multicol}

\usepackage{xcolor}
\PassOptionsToPackage{numbers}{natbib}
\definecolor{url}{RGB}{0,73,147}
\definecolor{mypink}{HTML}{bc4749}
\definecolor{mygray}{gray}{.9}

\newcommand{\cmark}{\ding{51}}%
\newcommand{\xmark}{\ding{55}}%

\usepackage{array}
\usepackage{colortbl}

\usepackage{pifont}    
\usepackage{bbding}       

\usepackage{caption}
\usepackage{enumitem}
\usepackage{multirow}
\makeatletter
\newcommand{\thickhline}{%
 \noalign {\ifnum 0=`}\fi \hrule height 1pt
 \futurelet \reserved@a \@xhline
}
\makeatother
\newcommand{\pub}[1]{\color{gray}{\tiny{[{#1}]}}}

\usepackage[accsupp]{axessibility}  

\title{MMAD: Multi-label Micro-Action Detection in Videos}

\author{Kun Li$^{1}$, Pengyu Liu$^{1}$, Dan Guo$^{1,2}$\thanks{Corresponding author}, Fei Wang$^{1,2}$, Zhiliang Wu$^{3}$, Hehe Fan$^{3}$, Meng Wang$^{1}$\footnotemark[1] \\
$^{1}$School of Computer Science and Information Engineering, Hefei University of Technology \\
$^{2}$Institute of Artificial Intelligence, Hefei Comprehensive National Science Center \\ 
$^{3}$ ReLER, CCAI, Zhejiang University \\
{\tt\small kunli.hfut@gmail.com}, 
{\tt\small guodan@hfut.edu.cn}, 
{\tt\small eric.mengwang@gmail.com}
}

\begin{document}
\maketitle

\begin{abstract}
Human body actions are an important form of non-verbal communication in social interactions. This paper specifically focuses on a subset of body actions known as micro-actions, which are subtle, low-intensity body movements with promising applications in human emotion analysis. 
In real-world scenarios, human micro-actions often temporally co-occur, with multiple micro-actions overlapping in time, such as concurrent head and hand movements. However, current research primarily focuses on recognizing individual micro-actions while overlooking their co-occurring nature. 
To address this gap, we propose a new task named \textbf{M}ulti-label \textbf{M}icro-\textbf{A}ction \textbf{D}etection (\textbf{MMAD}), which involves identifying all micro-actions in a given short video, determining their start and end times, and categorizing them. 
Accomplishing this requires a model capable of accurately capturing both long-term and short-term action relationships to detect multiple overlapping micro-actions. 
To facilitate the MMAD task, we introduce a new dataset named \textbf{M}ulti-label \textbf{M}icro-\textbf{A}ction-\textbf{52} (\textbf{MMA-52}) and propose a baseline method equipped with a dual-path spatial-temporal adapter to address the challenges of subtle visual change in \textbf{MMAD}. 
We hope that MMA-52 can stimulate research on micro-action analysis in videos and prompt the development of spatio-temporal modeling in human-centric video understanding.  
The proposed MMA-52 dataset is available at: \href{https://github.com/VUT-HFUT/Micro-Action}{https://github.com/VUT-HFUT/Micro-Action}.
\end{abstract}

\section{Introduction}~\label{sec:intro}
Human body actions, as an important form of non-verbal communication, effectively convey emotional information in social interactions~\cite{aviezer2012body}. 
Previous research primarily focused on interpreting classical expressive emotions through facial expressions~\cite{hinduja2020recognizing,wu2023patch,liu2024uncertain,dindar2020leaders}, speech~\cite{iyer2022comparison,liu2018unsupervised,zhao2025temporal}, or expressive body gestures~\cite{balazia2022bodily,liu2021imigue,chen2023smg,li2023joint,li2023data}. 
In contrast, our study shifts the focus to a specific subset of body actions known as Micro-Actions (MAs)~\cite{liu2021imigue,chen2023smg,liu2024micro,guo2024benchmarking,guo2024mac,gong2024micro,li2025prototypical}. MAs are imperceptible non-verbal behaviors characterized by low-intensity movement with potential applications in emotion analysis.

\begin{figure}[t!]
\centering
\includegraphics[width=1\linewidth]{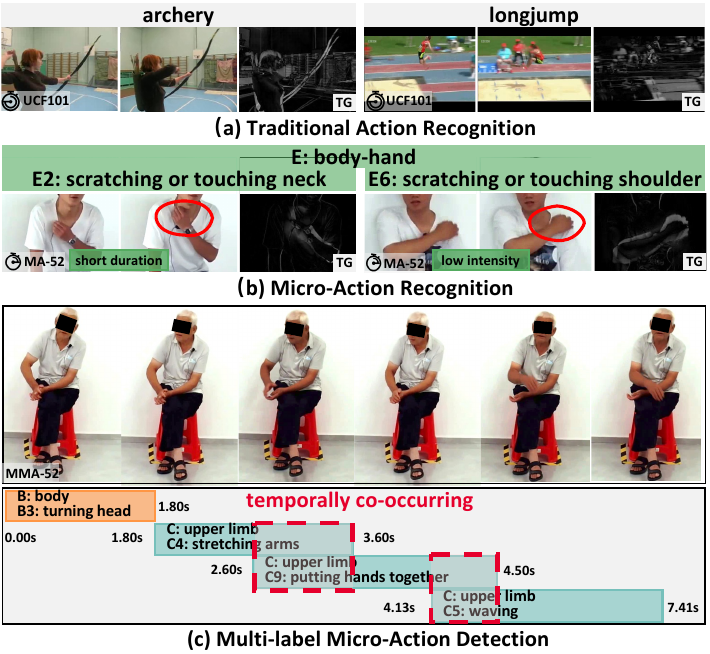}
\caption{(a) Traditional Action Recognition~\cite{soomro2012ucf101,kuehne2011hmdb,carreira2019short} primarily focuses on actions involving large and observable movements. (b) Micro-Action Recognition~\cite{liu2021imigue,guo2024benchmarking,li2025prototypical} targets fine-grained actions at \texttt{body-level} and \texttt{action-level}, characterized by short duration, low intensity, and minor difference. Temporal Gradient (TG)~\cite{xiao2021learning} is used to visualize the subtle changes. (c) \textbf{M}ulti-label \textbf{M}icro-\textbf{A}ction \textbf{D}etection (\textbf{MMAD}) aims to detect all micro-actions within a short video, accounting for temporal co-occurrence.}
\label{fig:intro}
\end{figure}

Compared to traditional actions~\cite{soomro2012ucf101,kuehne2011hmdb,carreira2019short,fan2018unsupervised,li2022dynamic,li2025repetitive,wang2025exploiting,fan2021point}, MAs have distinct characteristics as follows: \textbf{(1) Short duration.} 
As shown in Fig.~\ref{fig:intro}, MAs typically last only a few seconds, exhibiting subtle visual changes between consecutive frames. For instance, ``touching neck'' only exhibits minor changes in the neck region between a few frames. In contrast, conventional actions typically last around 5–10 seconds and involve larger and more dynamic motions in hundreds of frames. For example, the movements of ``archery'' or ``jump'' involve large motions. 
\textbf{(2) Low intensity.}
MAs are characterized by minor spatial distinctions.
As shown in Fig.~\ref{fig:intro} (b), the difference between ``touching neck'' and ``touching shoulder'' varies only in the specific contact regions involved. In contrast, conventional actions usually with identifiable motion patterns, such as those in ``longjump'' in Fig.~\ref{fig:intro} (a), where the overall movement is more visually distinct. 
\textbf{(3) Fine-grained categories.} MAs demand classification at both the body-part and action levels, involving isolated movements of individual body parts (\eg, ``head'', ``upper limb'', and ``lower limb'') as well as coordinated motions combining parts (\eg, ``head-hand,'' and ``body-hand''). In contrast, conventional action recognition typically focuses on larger-scale, whole-body motions. 

Remarkable progress has been made in the micro-action recognition~\cite{guo2024benchmarking,li2025prototypical,gong2024micro,wang2024instance,li2024advancing} task with the advancement of Vision Transformers~\cite{li2023vigt,wang2024frequency,wang2024eulermormer,wu2025bvinet,li2023datae,wu2024waveformer}. However, in the real world, micro-actions are \textbf{naturally temporal co-occurring}, which poses challenges for traditional action recognition methods.
As shown in Fig.~\ref{fig:intro} (c), different micro-actions may occur simultaneously, such as ``stretching arms'' frequently happening simultaneously with ``putting hands together''. 
Therefore, driven by this intuition, we propose a new task named Multi-label Micro-Action Detection (\textbf{MMAD}) that recognizes all the micro-actions in the video sequence, achieving a fine-grained understanding of micro-actions. 
MMAD involves identifying all micro-actions within the video and determining their corresponding start and end times, as well as their categories. 
\textbf{Firstly}, MMAD requires a model capable of capturing both long-term and short-term action relationships to locate multi-scale micro-actions. 
\textbf{Secondly}, the model also needs to explore the complex inter-relationships between different micro-actions to ensure comprehensive detection of all possible micro-actions. 
\textbf{Finally}, due to the inherent nature of short duration and subtle movements in micro-actions, there is also a greater challenge in recognizing the correct categories. 

To facilitate this research, we collect the first large-scale \textbf{M}ulti-label \textbf{M}icro-\textbf{A}ction-\textbf{52} (\textbf{MMA-52}) dataset, which consists of 6,528 ($\sim$6.5k) videos with 19,782 ($\sim$20k) action instances from 203 subjects. 
We first evaluate 10 baselines for traditional action detection on the MMA-52 dataset, including multi-label action detection methods and conventional temporal action detection methods. 
Next, we propose a baseline that incorporates a dual-path spatial-temporal adapter to capture the subtle visual changes between frames and model the associations between different actions. 
Specifically, the designed dual-path spatial-temporal adapter consists of two parts. In spatial, we use a depth-wise 2D convolution to model the subtle changes between adjacent frames. In temporal, we apply 1D temporal depth-wise convolution to aggregate temporal information. 
Finally, we use two learnable parameters to fuse temporal and spatial features separately. 
Extensive experiments and error analyses are conducted on the proposed benchmark dataset to validate the effectiveness of the proposed method. 

Overall, the main contributions of this paper are summarized as follows:
\begin{itemize}
\item We introduce the task of multi-label micro-action detection (MMAD) and collect the multi-label micro-action-52 (MMA-52) dataset to facilitate the research of micro-action analysis. 
\item We propose an initial solution with a dual-path spatio-temporal adapter to model subtle discriminative motions. Experimental results on the benchmark dataset validate the effectiveness of the proposed method. 
\item We evaluate 10 baselines from the conventional temporal action detection on MMAD and in-depth studies, which reveal the inherent challenges in multi-label micro-action detection.
\end{itemize}

\begin{figure*}[!t]
\centering
\includegraphics[width=1.0\linewidth]{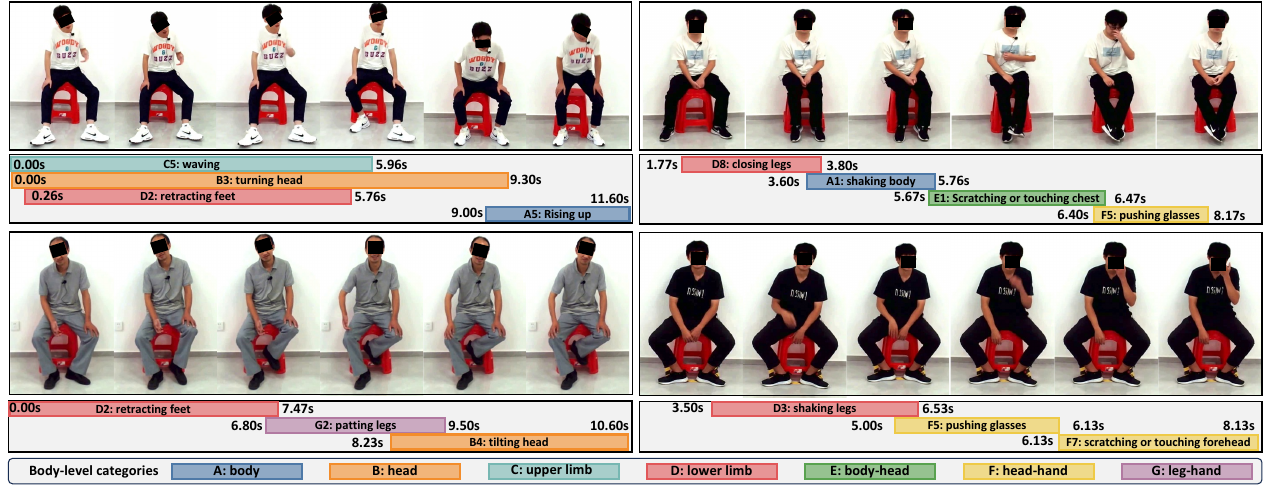}
\caption{
\textbf{Video samples from the MMA-52 dataset.} For each sample, there are different micro-action that occurs at the same time, increasing the challenge of identifying accurate micro-actions.  
}
\label{fig:demo_mma52}
\end{figure*}

\section{Related Work}
\subsection{Micro-Actions Recognition}
Micro-Actions (MAs)~\cite{liu2021imigue,chen2023smg,guo2024benchmarking,chen2024prototype,li2025prototypical} are an important form of non-verbal communication, which are usually related to humans’ emotional status~\cite{aviezer2012body}. To facilitate the study of these subtle movements, several datasets have been constructed. 
To advance the study of these subtle movements, several datasets have been developed. 
iMiGUE~\cite{liu2021imigue} and SMG~\cite{chen2023smg} focused on spontaneous micro-gestures in the upper limbs of athletes, revealing deep emotional states conveyed through these micro-gestures. 
In contrast, MPIIGI~\cite{balazia2022bodily} primarily examined subtle upper-body behaviors in group interactions. 
To better analyze and understand the whole-body movement, Guo~\etal~\cite{guo2024benchmarking} proposed a large-scale micro-action dataset named Micro-Action (MA-52), which consists of 52 action-level MAs within 7 body-level in whole-body. They also evaluated conventional action recognition methods, including 2D CNN based~\cite{wang2018temporal,lin2019tsm,feichtenhofer2019slowfast}, 3D CNN based~\cite{C3D,I3D}, GCN-based~\cite{yan2018spatial,liu2020disentangling}, and Transformer-based~\cite{liu2022video}. 
More recently, Li~\etal~\cite{li2025prototypical} proposed a prototypical calibrating ambiguous network, designed to mitigate the influence of the inherent ambiguity of micro-actions in micro-action recognition. 

\subsection{Temporal Action Detection} 
Temporal action detection (TAD)~\cite{dai2022ms,tan2022pointtad,lin2019bmn,lin2018bsn,liu2024end} aims to localize and classify actions in untrimmed video sequences. There have been many benchmarks focused on different domains, such as sports (THUMOS14~\cite{THUMOS14} and FineGym~\cite{shao2020finegym}), kitchen activities
(MPII Cooking~\cite{rohrbach2012database} and EPIC-Kitchens~\cite{rohrbach2016recognizing}), and daily events (ActivityNet~\cite{caba2015activitynet}, HACS Segment~\cite{zhao2019hacs}, and FineAction~\cite{liu2022fineaction}). Driven by these datasets, TAD has witnessed significant progress, leading to the emergence of advanced methods. These approaches can be broadly categorized into feature-based methods~\cite{dai2022ms,zhang2022actionformer,lin2018bsn,lin2019bmn,lin2020fast,xu2020g}, which rely on pre-extracted features to detect actions, and end-to-end learning-based methods~\cite{wang2021rgb,tan2022pointtad,liu2022empirical,liu2024end,lin2021learning}, which directly process raw video inputs for action localization and classification.
\textit{However}, micro-action detection is still in its infancy due to the lack of large-scale datasets. Micro-action detection remains in its early stages due to the absence of large-scale datasets. The most relevant datasets for our research are iMiGUE~\cite{liu2021imigue} and SMG~\cite{chen2023smg}, which focus on upper limb micro-gesture detection. 
Unfortunately, these datasets are not publicly accessible due to privacy issues. 
Compared to these datasets, MMA-52 surpasses existing benchmarks in terms of category diversity, number of subjects, and action instances. MMA-52 also features hierarchical labels, enabling more precise identification of multi-level MAs. We hope our MMA-52 can facilitate the research community to build robust algorithms for micro-action detection.

\section{The MMA-52 dataset}
\subsection{Dataset Construction}
\textbf{Data Collection.} 
The proposed Multi-label Micro-Action-52 (MMA-52) dataset is built upon the MA-52-Pro dataset collected by~\cite{guo2024benchmarking}. 
However, MA-52-Pro is not directly applicable to micro-action detection tasks due to the following reasons: \textbf{\textit{1) Lack of fine-grained annotations}}: MA-52-Pro does not provide detailed start/end timestamps for individual action instances. \textbf{\textit{2) Variation in video lengths}}: The videos in MA-52-Pro vary significantly in length, each video contains 1 to 15 MAs with durations ranging from 5s to beyond 100s. Such substantial imbalance in the action instances makes it unsuitable for action detection tasks. 
To address these challenges, \textbf{first}, we segment the videos into smaller clips, each ranging from 5 to 15 seconds, to ensure more consistency in video sequence and action instances. \textbf{Next}, we annotate each micro-action instance with its corresponding categories and precise start/end timestamps.

\begin{figure*}[t!]
\centering
\includegraphics[width=1.0\linewidth]{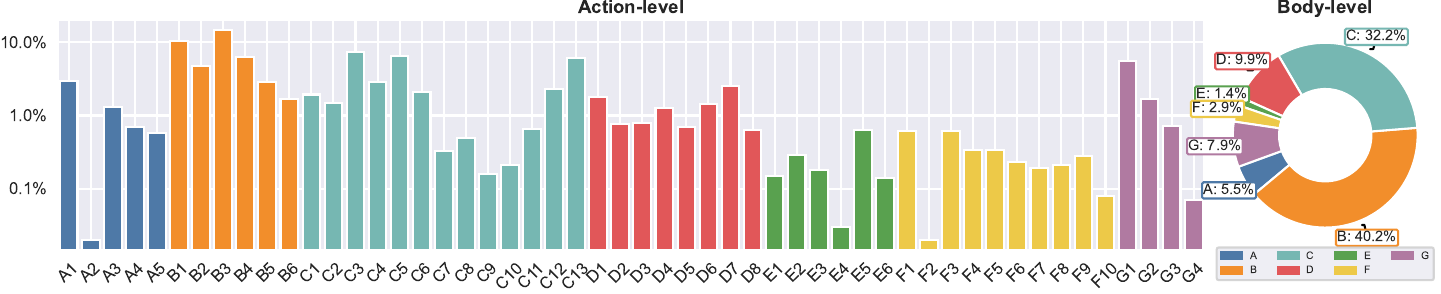}
\caption{
\textbf{The ratio of each action instance category in the MMA-52 dataset.} ``A1, A2, \ldots, G4'' denote the \texttt{action-level} categories while ``A, B, \ldots, G'' denote the \texttt{body-level} categories. The distribution of micro-action instances across the \texttt{action-level} and \texttt{body-level} follows a long-tail pattern. The definition of each category is the same as that of the MA-52 dataset~\cite{guo2024benchmarking}.}
\label{fig:class_static}
\end{figure*}

\begin{figure*}[t!]
\centering
\includegraphics[width=1.0\linewidth]{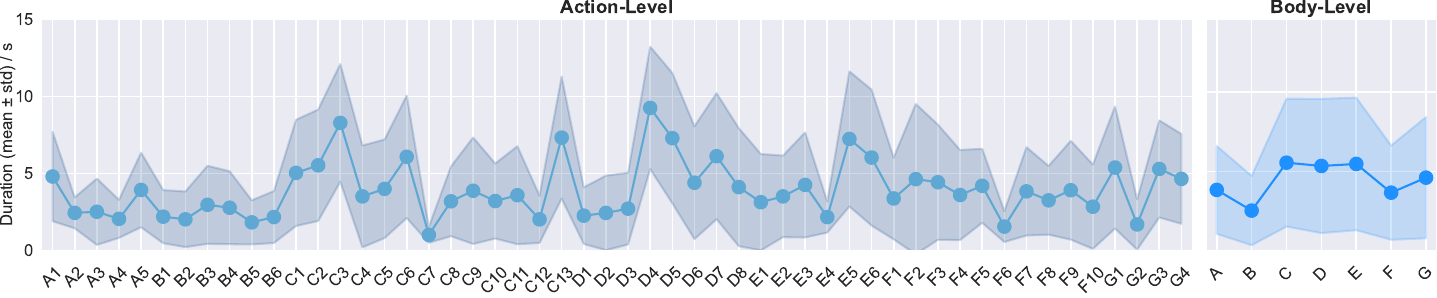}
\caption{
\textbf{The duration of each action instance in the MMA-52 dataset.} At the \texttt{action-level}, the duration of different actions varies significantly, with certain actions (\eg, C3, D4, E6) exhibiting notably longer durations, while others remain considerably shorter. In contrast, the \texttt{body-level} shows a relatively lower standard deviation, indicating that body-level labels tend to be more stable.
}
\label{fig:duration_static}
\end{figure*}

\noindent\textbf{Data Annotation.} Considering the inherent hierarchical nature of micro-actions~\cite{guo2024benchmarking,li2025prototypical}, each micro-action instance are annotated with \textit{body-level} and \textit{action-level} labels. 
In practice, annotating the multi-label micro-actions was a challenging and time-consuming task, as different types of MAs can occur simultaneously at any given moment, as illustrated in Fig.~\ref{fig:intro} (c).  
To ensure the accuracy of these annotations, we implemented three key measures to maintain the quality of the dataset. 
\textbf{\textit{1) Annotator training}}: Given the diversity of micro-action categories and the subtle differences between actions, we began by training the annotators.  
They first gained a thorough understanding of the definitions of the micro-action categories. 
Following this, we randomly selected 50 samples from each category in the micro-action recognition dataset (MA-52~\cite{guo2024benchmarking}) and asked the annotators to perform trial annotations. Feedback based on reference annotations was provided to correct any errors arising from misunderstandings. This step ensured the annotators were well-prepared before starting the actual annotation task. 
\textbf{\textit{2) Individual annotations}}: Each video segment is labeled independently by three trained annotators. For each action instance, both \texttt{body-level} and \texttt{action-level} categories are assigned, along with the corresponding start and end times. 
\textbf{\textit{3) Cross-check}}: After the individual annotation is completed, a cross-check is performed. If the temporal intersection-over-union (tIoU) of the same action instances annotated by all three annotators is greater than 0.9, the annotation is considered reliable. For any inconsistencies, the three annotators will discuss and follow the majority decision as the final result. This process ensures the accuracy and consistency of the annotations. 

\noindent\textbf{Data Partition.} Since the same micro-action may exhibit individual differences, we utilize a subject-independent data partitioning strategy. As shown in Table~\ref{tab:sta_mma52}, the training, validation, and test sets consist of different individuals, ensuring that no individual appears in more than one set. This strategy helps to better evaluate the model's performance across diverse subjects and enhances its generalization ability on unseen data.

\begin{table}[t!]
\centering
\tabcolsep 2pt
\caption{\textbf{Data statistics for the MMA-52 dataset.} ``Duration'' refers to the length of all videos, ``Avg. Video'' denotes the average length of videos, and ``Avg. Insta.'' represents the average length of action instances. ``\#Subj.'' denotes the number of subjects.}
\vspace{-0.3em}
\resizebox{1.0\linewidth}{!}{
\begin{tabular}{c|ccccc|c}
\hline\thickhline
\rowcolor{mygray}
Split & Videos & Instances & Duration & Avg. Video & Avg. Insta. & \#Subj. \\ \hline
Training & 4,534 &  13,698 & 12.91h & 10.25s & 4.10s & 140 \\
Validation & 1,475 & 4,735 & 4.34h & 10.60s & 4.10s & 37  \\
Test & 519 & 1,349 & 1.42h & 9.86s & 3.79s & 26  \\ \hline
\rowcolor[HTML]{f8f9fa}
All & 6,528 & 19,782 & 18.67h & 10.30s & 4.07s & 203 \\ 
\hline
\end{tabular}}
\label{tab:sta_mma52}
\end{table}

\subsection{Dataset Statistics and Properties} 
Table~\ref{tab:sta_mma52} presents the data statistics of the MMA-52 dataset, which consists of 6,528 videos, each ranging from 5 to 15 seconds in duration. The dataset contains a total of 19,782 full-body micro-action instances across 52 distinct action categories. On average, each video includes 3.1 action instances, with each instance lasting approximately 4.07 seconds. 
Fig.~\ref{fig:class_static} presents the proportion of each action category across the three subsets. 
The MMA-52 dataset includes long and short micro-actions with varying temporal durations and transitions between different action states. As shown in Fig.~\ref{fig:demo_mma52}, we illustrate some video samples from the MMA-52 dataset. 

Based on the above data statistics, the characteristics of the proposed MMA-52 dataset can be summarized as follows.
\textbf{\textit{1) Long-tail category distribution.}} As shown in Fig.~\ref{fig:class_static}, the dataset exhibits a long-tail distribution, where some MA occur frequently while others rarely occur. For example, ``B3: Turning head'' accounts for nearly 13\% whereas ``A2: Turning around'' appears not up to 1\%. This imbalance presents a challenge for models to enhance their robustness and ability to generalize in rare MAs. 
\textbf{2) \textit{Variability in micro-action hierarchies.}} As shown in Fig.~\ref{fig:duration_static}, micro-action duration exhibits significant variability, particularly at the action level, where some actions (\eg, C3, D4, E6) last considerably longer than others. The high standard deviation suggests substantial fluctuations in duration across individuals and scenarios, posing challenges for model learning and prediction. In contrast, the body-level shows a relatively lower standard deviation, indicating that body-level labels are more stable. This highlights the need for models to capture hierarchical relationships, as individual actions are influenced by different body parts, making it insufficient to rely solely on global features. 
\textbf{\textit{3) Subject-independent evaluation.}} Given the subtle nature of micro-actions, and micro-action patterns will be different across individuals, the MMA-52 dataset adopts a subject-independent setting. This means that the dataset includes action instances from a diverse range of subjects, ensuring that there is no overlap of subjects between the training, validation, and test sets. 
The goal is to ensure that the model learns to identify and generalize micro-actions across varying body types and movements.

\begin{table}[t!]
\centering
\tabcolsep 2pt
\caption{\textbf{Comparison with related datasets in micro-action analysis.} ``\#C'' denotes the number of categories. ``\#Subj.'' denotes the number of subjects. ``H-L'' denotes the hierarchy action label.}
\resizebox{1.0\linewidth}{!}{
\begin{tabular}{c|cccccc|c|c}
\hline\thickhline
\rowcolor{mygray}
Dataset & \#C & \#Subj. & H-L & Video & Instance & Duration & Task & Public \\ \hline
PAVIS F-T~\cite{beyan2020analysis} & 2 & 64 & \xmark &  64 & N/A & N/A & Recognition & \xmark \\ 
MPIIGI~\cite{balazia2022bodily} & 15 & 78 & \xmark & 7,905 & 7,905 & 2.13s & Recognition & \cmark \\
iMiGUE~\cite{liu2021imigue} & 32 & 72 & \xmark & 359 & 18,499 & 2.55s & Recognition & \xmark \\
SMG~\cite{chen2023smg} & 16 & 40 & \xmark & 414 & 3,712 & 2.14s & Recognition & \xmark \\ 
MA-52~\cite{guo2024benchmarking} &  52 & 205 & \cmark & 22,422 & 22,422 & 1.97s & Recognition & \cmark \\
\hline
iMiGUE~\cite{liu2021imigue} & 32 & 72 & \xmark &359 & 18,499 & 2.55s & Detection & \xmark \\
SMG~\cite{chen2023smg} & 16 & 40 &\xmark & 414 & 3,712 & 2.14s & Detection & \xmark \\ 
\rowcolor[HTML]{f8f9fa}
\textbf{MMA-52 (Ours)} & \textbf{52} & \textbf{203} & \cmark & \textbf{6,528} & \textbf{19,782}  & 4.07s & Detection & \cmark  \\ \hline
\end{tabular}}
\label{tab:compare}
\end{table}

\subsection{Comparison with Existing Datasets}
We first review the related datasets~\cite{beyan2020analysis,balazia2022bodily,liu2021imigue,chen2023smg} in micro-action recognition. 
PAVIS F-T~\cite{beyan2020analysis} and MPIIGI~\cite{balazia2022bodily} focus on body behaviors in group social interactions. The former concentrates solely on face-touching versus non-face-touching, while the latter analyzes a broader range of behavior categories (\eg, scratch and shrug). In contrast, iMiGUE~\cite{liu2021imigue} and SMG~\cite{chen2023smg} target upper limb micro-gestures. 
However, these datasets are relatively limited in terms of category diversity and subjects. To address this limitation, MA-52~\cite{guo2024benchmarking} introduces a large-scale micro-action recognition dataset with 52 categories. 
Then, we compare with micro-action detection datasets. iMiGUE and SMG can be applied to action detection tasks, but they are limited to action categories. 
In contrast, our MMA-52 benefits from the hierarchy label (\ie, \texttt{body-level} and \texttt{action-level}), large-scale action instances involving diverse subjects, enabling the design of more comprehensive and scalable detection models. 

\section{Methodology}~\label{sec:method}
\subsection{Problem Formulation}
The Multi-label Micro-Action Detection (MMAD) can be formulated as a set prediction problem of micro-action instances. 
Let the video be $V$ containing $T$ frames. 
The annotation of video $V$ is composed by a set of micro-action instances $\Psi = \{\varphi=(t_n^s, t_n^e, c_n)\}^{N_{g}}_{N=1}$, where $N_{g}$ is the number of micro-action instances, $t_n^s$ and $t_n^e$ are the starting and ending timestamp of the $n$-th micro-action instance, $c_n$ is its micro-action category. 
The model $\gF$ is required to predict a set of micro-action proposals $\hat{\Psi} = \{\hat{\varphi}=(\hat{t_n^s}, \hat{t_n^e}, \hat{c_n})\}^{N_{p}}_{N=1}$, where $\hat{t_n^s}$ and $\hat{t_n^e}$ are the predicted starting and ending timestamp of the $n$-th micro-action instance, $\hat{c_n}$ is its predicted micro-action category. $N_p$ is the number of predicted micro-action instances.

\subsection{Preliminary}
Before introducing the baseline, we first briefly review the related techniques used in this paper.

\begin{figure}[t!]
\centering
\includegraphics[width=1\linewidth]{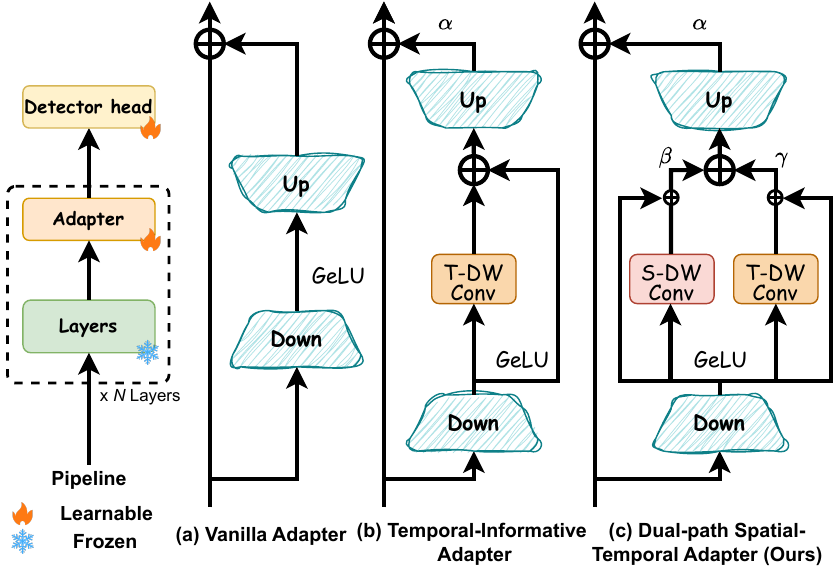}
\caption{\textbf{Action detection pipeline and architecture of different adapters.} (a) The vanilla adapter~\cite{houlsby2019parameter} adopts the bottleneck structure. (b) The baseline model~\cite{liu2024end} designed a temporal-informative adapter to aggregate temporal information. (c) The proposed dual-path spatial-temporal adapter aggregates both spatial and temporal information.}
\label{fig:archs_adaper}
\end{figure}

\noindent\textbf{Vanilla Adapter.} 
As illustrated in Fig.~\ref{fig:archs_adaper} (a), the vanilla adapter~\cite{houlsby2019parameter} comprises a down-projection and an up-projection fully connected (FC) layer, with a non-linear activation function $\sigmoid$ (such as GeLU~\cite{hendrycks2016gaussian}) applied between the two projections. Subsequently, a residual connection is applied to the output of the projection layer. This process can be formulated as follows:
\begin{equation}
\mX^{\prime}=\texttt{Adapter}(\mX)= \mW_{up}^{\top} \cdot \sigmoid (\mW_{down}^{\top}\cdot\mX) + \mX, 
\end{equation}
where $\mW_{down}\in\R^{d\times\frac{d}{r}}$ and $\mW_{up}\in\R^{\frac{d}{r}\times d}$ denote the parameter of down- and up-projection, respectively. $r$ is a downsampling ratio greater than one. 

\noindent\textbf{AdaTAD}~\cite{liu2024end}. 
As shown on the left in Fig.~\ref{fig:archs_adaper}, there is the pipeline of AdaTAD (Adapter fine-tuning for temporal action detection)~\cite{liu2025opentad}. An adapter is inserted into the backbone layers (\eg, VideoMAE~\cite{tong2022videomae}) to fine-tune the model for action detection. 
Since the standard adapter is limited to adapting channel information, AdaTAD proposes a Temporal-Information Adapter (TIA) designed to aggregate informative local context from neighboring frames, enhancing temporal action detection.
As shown in Fig.~\ref{fig:archs_adaper} (b), the temporal-informative adapter inserts a temporal depth-wise convolution (denoted as \texttt{T-DWConv}) with the kernel size of $k\times 1\times 1$ and group size of $r$ for depth-wise convolution to model the local contexts between adjacent frames. The TIA module is inserted into the layers between different backbones to realize transfer learning for efficient parameter tuning. Overall, the TIA module can be formulated as follows:
\begin{equation}
\begin{aligned}
\mX^{\prime} &= \texttt{TIA}(\mX) \Leftrightarrow\\
&\left\{\begin{array}{l}
\bar{\mX} = \sigmoid (\mW_{down}^{\top}\cdot\mX), \\
\hat{\mX} = \underbrace{\texttt{T-DWConv}(\hat{\mX})}_{\text{{\color{brown}\textbf{temporal contexts}}}} + \bar{\mX}, \\
\mX^{\prime} = \alpha \cdot \mW_{up}^{\top} \cdot \hat{\mX} + \mX,
\end{array}\right.
\end{aligned}
\end{equation}
where $\alpha$ is a learnable parameter.

\input{tables/main_results}

\subsection{Dual-path Spatial-Temporal Adapter}
As stated in the introduction, micro-action involves the subtle visual difference between adjacent frames and short duration. The baseline AdaTAD only utilizes the temporal depth-wise convolution layers to aggregate temporal information, but we argue that this will limit the model from capturing the discriminative spatial features of micro-actions. Therefore, we propose a simple Dual-path Spatial-Temporal Adapter (DSTA) to model spatial changes and temporal correlations separately. Then, the learned features from the spatial path and temporal path are fused based on the learnable weights. The above process can be formulated as follows: 
\begin{equation}
\begin{aligned}
\mX^{\prime} &= \texttt{DSTA}(\mX) \Leftrightarrow\\
&\left\{\begin{array}{l}
\bar{\mX} = \sigmoid (\mW_{down}^{\top}\cdot\mX), \\
\hat{\mX}_{s} = \underbrace{\texttt{S-DWConv}(\hat{\mX})}_{\text{{\color{pink}\textbf{spatial contexts}}}} + \bar{\mX}, \\
\hat{\mX}_{t} = \underbrace{\texttt{T-DWConv}(\hat{\mX})}_{\text{{\color{brown}\textbf{temporal contexts}}}} + \bar{\mX}, \\
\mX^{\prime} = \alpha \cdot \mW_{up}^{\top} \cdot [\beta \cdot \hat{\mX}_{s}; \gamma\cdot \hat{\mX}_{t}] + \mX,
\end{array}\right.
\end{aligned}
\end{equation}
where $[;]$ denotes the concatenation operation, $\beta$ and $\gamma$ are two learnable parameters to balance the weights of spatial and temporal pathways, respectively. \texttt{S-DWConv} symbols the spatial depth-wise convolution with the kernel size of $1\times 1$.

\section{Experiments}~\label{sec:exp}
\subsection{Experiments Setup}
\noindent\textbf{Evaluation Metrics.} 
We use the mean Average Precision (mAP)~\cite{tan2022pointtad,zhang2022actionformer,liu2024end} to evaluate the performance of multi-label micro-action detection. 
mAP measures the completeness of predicted action instances.
We report the average mAP results for tIoU thresholds ranging from 0.1 to 0.9 in increments of 0.1 and the average mAP at each specific threshold. 
Considering the hierarchy of micro-actions, we report Detection-mAP at both \texttt{body-level} and \texttt{action-level.} Additionally, we also report the average value of mAP (\texttt{AVG}) of these two levels. 

\noindent\textbf{Implementation Details.}
We conduct experiments with the open-source toolbox OpenTAD~\cite{liu2025opentad}. The model employs mixed-precision training and activation checkpointing to reduce memory usage.
Following~\cite{liu2024end}, we use ActionFormer~\cite{zhang2022actionformer} as the detector head, retaining the original hyperparameter settings. The backbone's learning rate remains fixed, while the adapter's learning rate varies between 1e-4 and 4e-4. 
By default, frame resolution is set to 160$^2$. 
For the body-level prediction, we follow the common practice in micro-action analysis~\cite{guo2024benchmarking,gu2025motion} by converting the action-level results to the body-level.

\subsection{Main Results}
Since Multi-label Micro-Action Detection (MMAD) is a new task, we evaluate 10 baselines in conventional temporal action detection.
The results are reported in Table~\ref{tab:main_results}. 
\textbf{(1)} For the multi-label temporal action detection methods (MS-TCT~\cite{dai2022ms} and PointTAD~\cite{tan2022pointtad}), these methods achieve the lowest performance. We attribute the significant performance drop to the gap between conventional actions and micro-actions.
\textbf{(2)} For the feature-based methods, TemporalMaxer~\cite{tang2023temporalmaxer} get the best average mAP of 20.34, while the TriDet~\cite{shi2023tridet} only achieves the average mAP of 16.04. 
\textbf{(3)} For the feature-based methods, the baseline method AdaTAD~\cite{liu2025opentad} exhibits better performance with the backbone model enlarged from the VideoMAE small version to the base version. The best result in average mAP is 21.80, which is better than all feature-based methods.  
Compared to the baseline, we can see that the proposed method exhibits consistent improvement on different backbones. Specifically, on the VideoMAE-S and VideoMAE-B, there are 1.63\%, and 2.26\% improvements on average mAP, respectively.  
Although the proposed method achieves the highest average mAP of 25.34, it is still far away from conventional action detection~\cite{liu2024end}.
These results indicate that there is still a gap in accurately identifying the micro-actions. 

\begin{table}[t!]
\centering
\caption{\textbf{Performance comparison under different adapters.}}
\resizebox{0.95\linewidth}{!}{
\begin{tabular}{l|ccc}
\hline\thickhline
\rowcolor{mygray}
Setting & Param. & mAP & gains \\ \hline
Snippet Feature &  - & 10.60 & - \\
+ Full fine-tuning & 20.89M & 15.40 & +4.80 \\
+ Standard Adapter~\cite{houlsby2019parameter} & 0.85M & 15.87 & +5.27 \\
+ TIA (AdaTAD~\cite{liu2024end}) & 0.96M &  16.25 & +5.65  \\ \hline
+ \textbf{DTSA (Ours)} & 1.80M & \textbf{18.16} & \textbf{+7.56} \\ \hline
\end{tabular}}
\vspace{-0.3em}
\label{tab:abl_adapter}
\end{table}

\subsection{Ablation Studies}

\noindent \textbf{The ablation of the adapters.} 
To validate the effectiveness of the proposed dual-path spatio-temporal adapter, we conduct experiments as follows: ``Full fine-tuning'' denotes fine-tuning the backbone, Standard Adapter~\cite{houlsby2019parameter}, and 
``TIA'' from the baseline model AdaTAD. 
The results are reported in Table~\ref{tab:abl_adapter}. 
The baseline model only achieves the mAP of 16.25\%, due to the neglect of crucial spatial information in micro-actions. In contrast, the proposed DSTA achieves the best results of 18.16 in terms of mAP, and there is 1.91\% improvement. 

\subsection{Error Analysis}
Following the convention practice~\cite{liu2024end,zhao2023re2tal,shi2023tridet,zhang2022actionformer} in action detection, we use the tool~\cite{alwassel2018diagnosing} to analyze the results.

\begin{figure}[t!]
\centering
\begin{subfigure}{\linewidth}
\centering
\includegraphics[width=\linewidth]{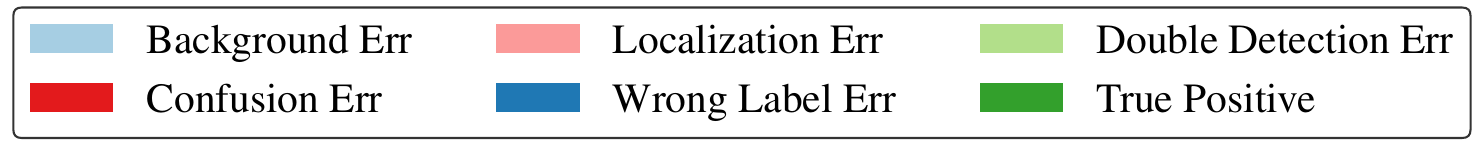}
\end{subfigure}
\begin{subfigure}{0.5\linewidth}
\centering
\includegraphics[width=\linewidth]{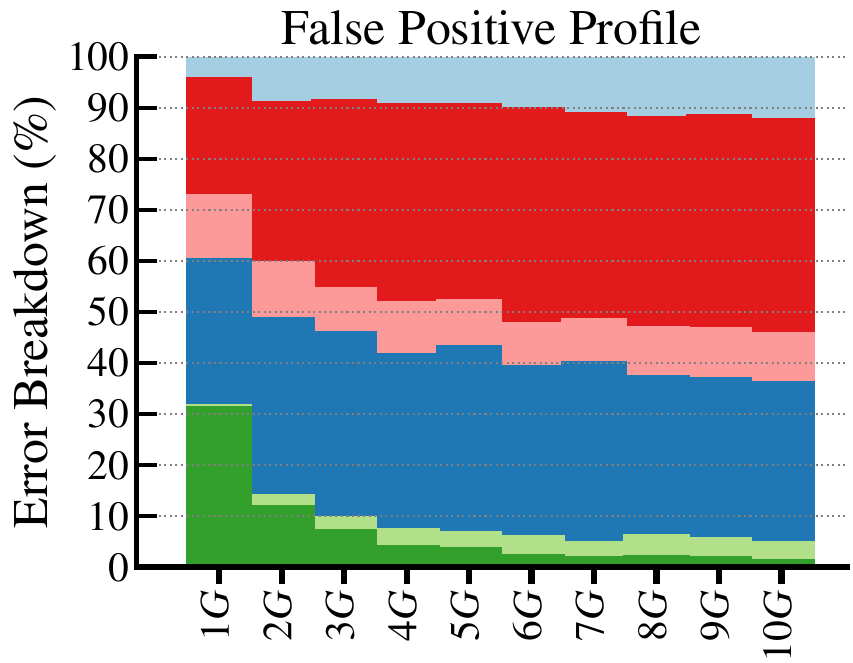}
\caption{\texttt{Action-level} Results}
\end{subfigure}%
\begin{subfigure}{0.5\linewidth}
\centering
\includegraphics[width=\linewidth]{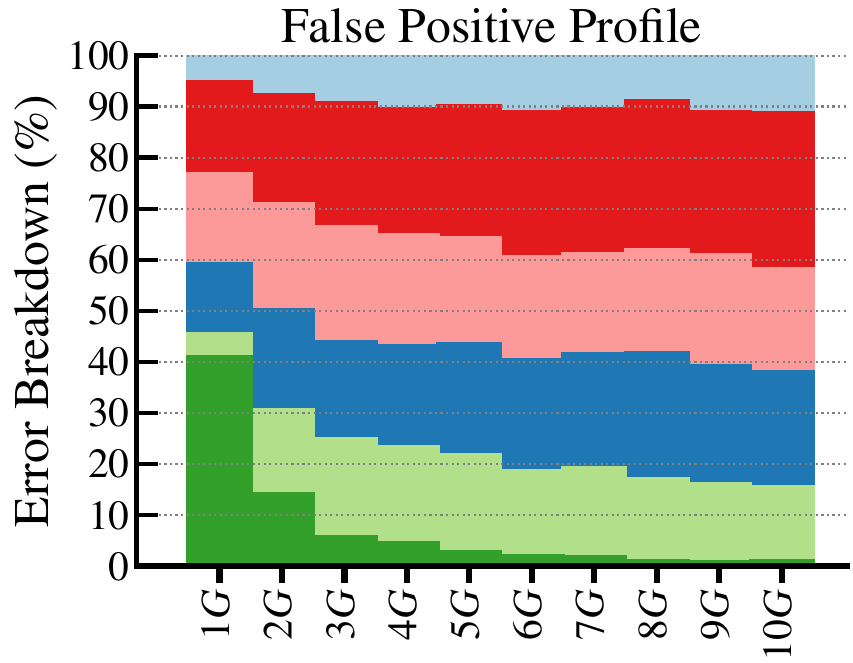}
\caption{\texttt{Body-level} Results}
\end{subfigure}
\vspace{-1.5em}  
\caption{\textbf{False Positive Profiling} on the MMA-52 dataset. 
The errors at the action-level mainly involve confusion and wrong label assignments while the body-level errors are evenly distributed.
}
\label{fig:fpa_ana}
\end{figure}

\noindent\textbf{False Positive Profiling.} 
As illustrated in Fig.~\ref{fig:fpa_ana}, we conduct false positive profiling at tIoU=0.5. The x-axis is top-$G$ predictions at tIoU=0.5, where $G$ refers to the number of ground-truth instances. 
In the \texttt{action-level}, false positive errors are mainly concentrated in background, localization, and label errors, with background errors being the most prominent. In contrast, at the \texttt{body-level}, the proportion of correct detections (True Positive) increases significantly, and the error distribution is more balanced, showing a more stable performance. 
In summary, false positive errors at the \texttt{action-level} indicate uncertainty in the model’s understanding of action-level MAs. The \texttt{body-level} shows higher accuracy, meaning the model can more accurately capture body movements.

\begin{figure}[t!]
\centering
\begin{subfigure}{\linewidth}
\centering
\includegraphics[width=0.8\linewidth]{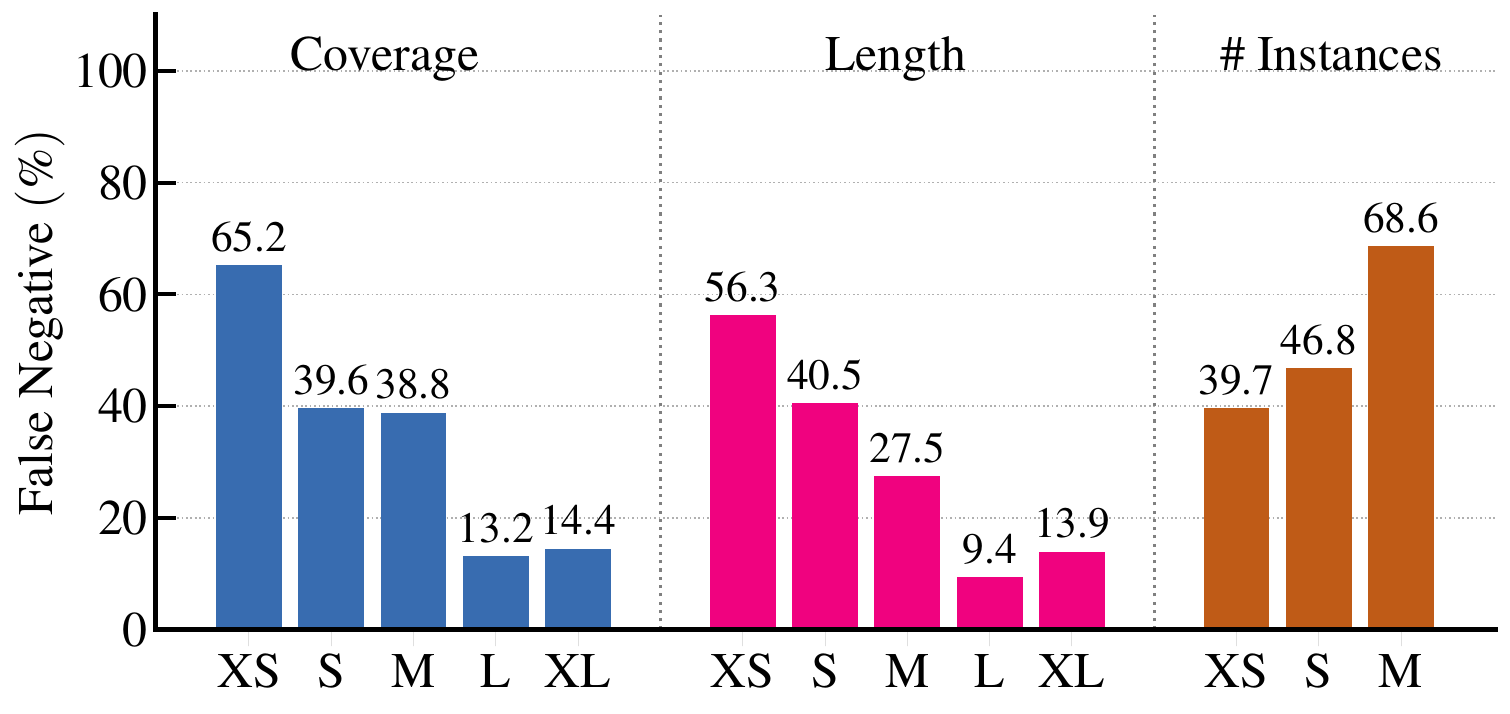}
\vspace{-0.1cm}  
\caption{\texttt{Action-level} Results}
\end{subfigure}
\vspace{-0.1cm}  
\begin{subfigure}{\linewidth}
\centering
\includegraphics[width=0.8\linewidth]{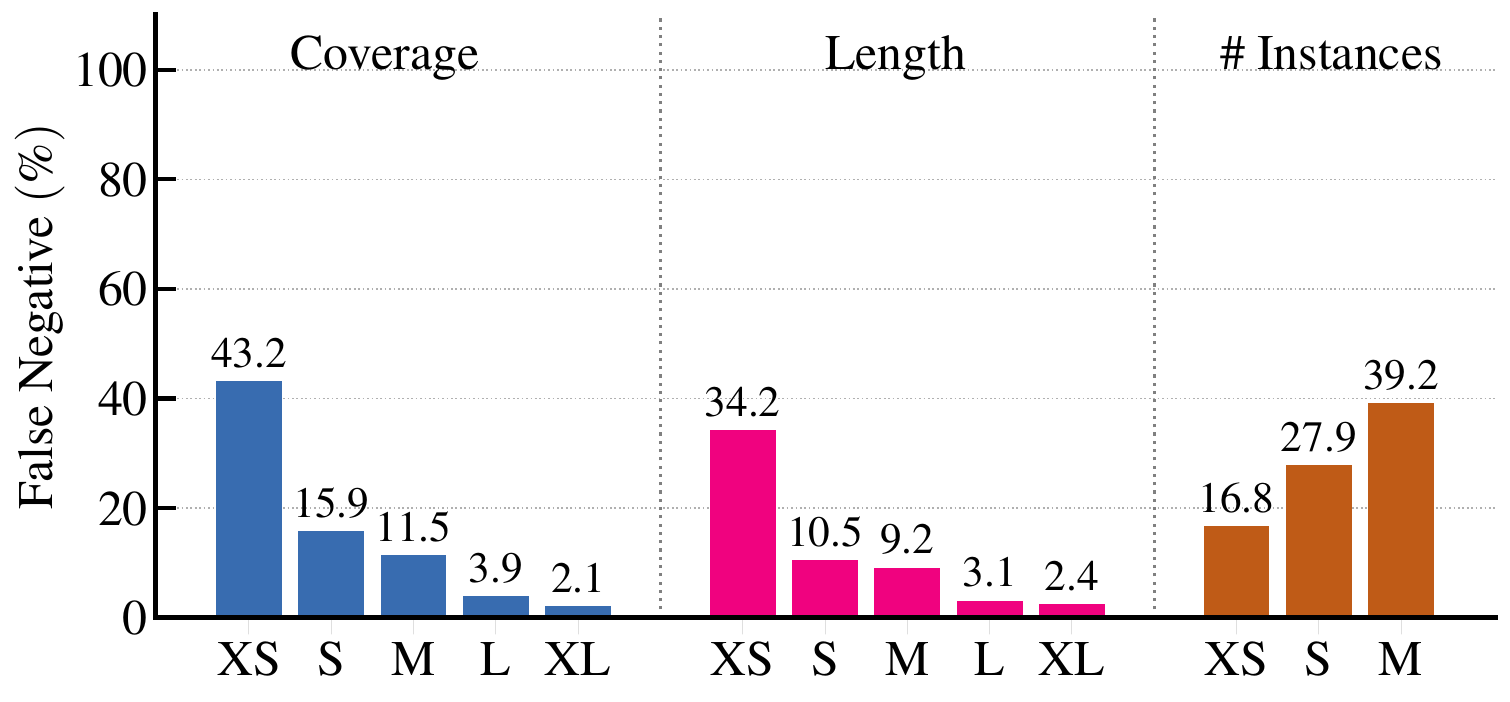}
\caption{\texttt{Body-level} Results}
\end{subfigure}
\caption{\textbf{False Negative Profiling} on the MMA-52 dataset. 
The false negatives are primarily found in instances with short coverage and duration.}
\label{fig:fna}
\end{figure}

\noindent\textbf{False Negative Profiling.}
As shown in Fig.~\ref{fig:fna}, we also conduct the False Negative Profiling under different characteristics. Specifically, ``Coverage'' denotes the ratio of instances within the video,  ``Length'' represents the duration (seconds) of instances, and ``\#Instances'' is the number of instances. 
Taking into account the statistics of the MMA-52 dataset, the range of these characteristics are as follows, \ie, ``Coverage'' refers to [0.0, 0.2, 0.4, 0.6, 0.8, 1.0], ``Length'' refers to [0, 3, 5, 8, 9, INF], and ``\#Instances'' refers to [-1, 2, 5, 7, INF]. These characteristic buckets are labeled as [XS, S, M, L, XL] on the axis. 
Based on the dimensions of Coverage and Length, we can see that False Negative rates are very high in the XS, S, and M buckets at the action-level, while only in the XS bucket at the body-level. These results suggest that detecting shorter micro-action instances remains a challenge, while longer instances are handled more effectively. 
In the dimension of ``\#Instances'', False Negative rates are primarily observed in the M bucket, indicating that the key challenge is in videos with dense instances. 
Overall, the future direction for MMAD should focus on improving the detection of instances with low coverage and short length.

\begin{figure}[t!]
\centering
\includegraphics[width=0.95\linewidth]{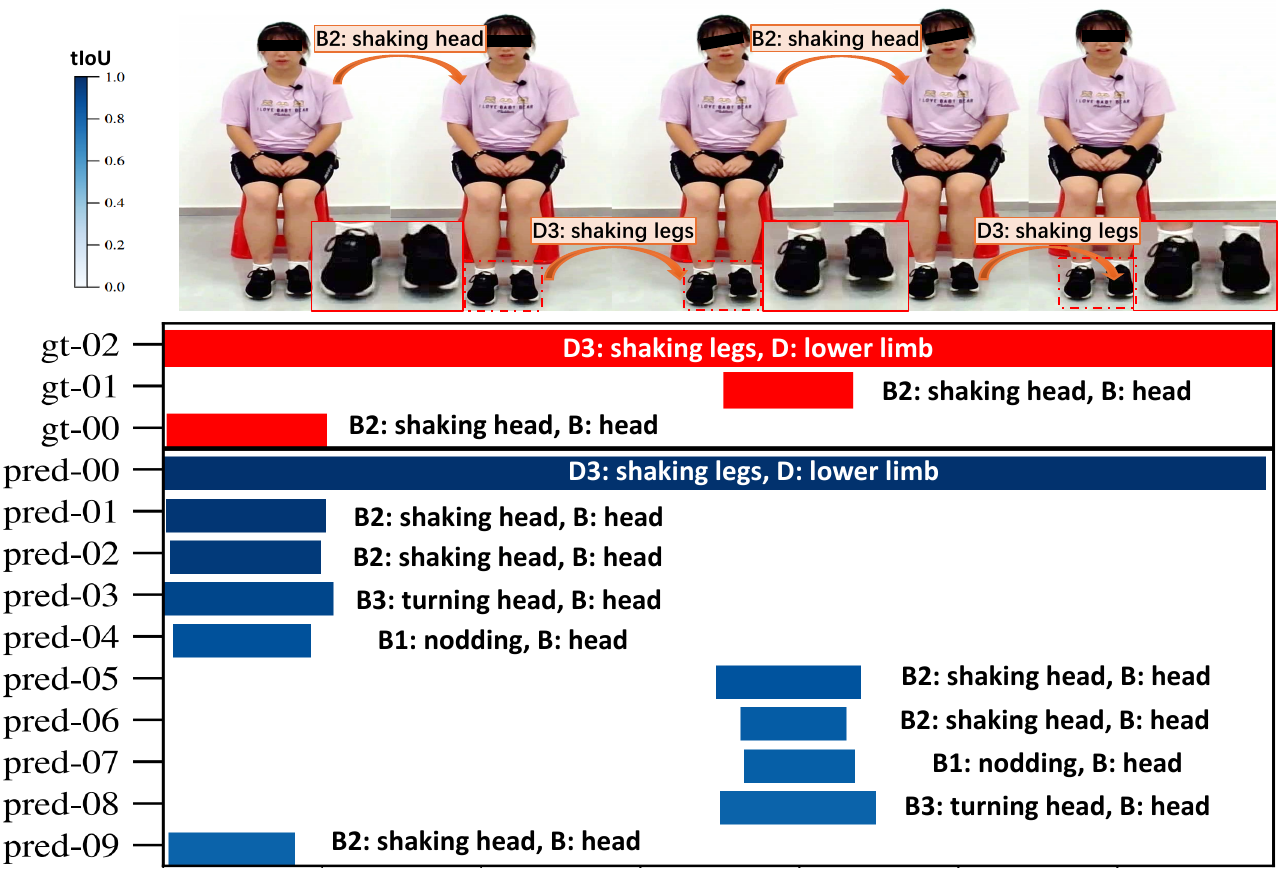}
\includegraphics[width=0.95\linewidth]{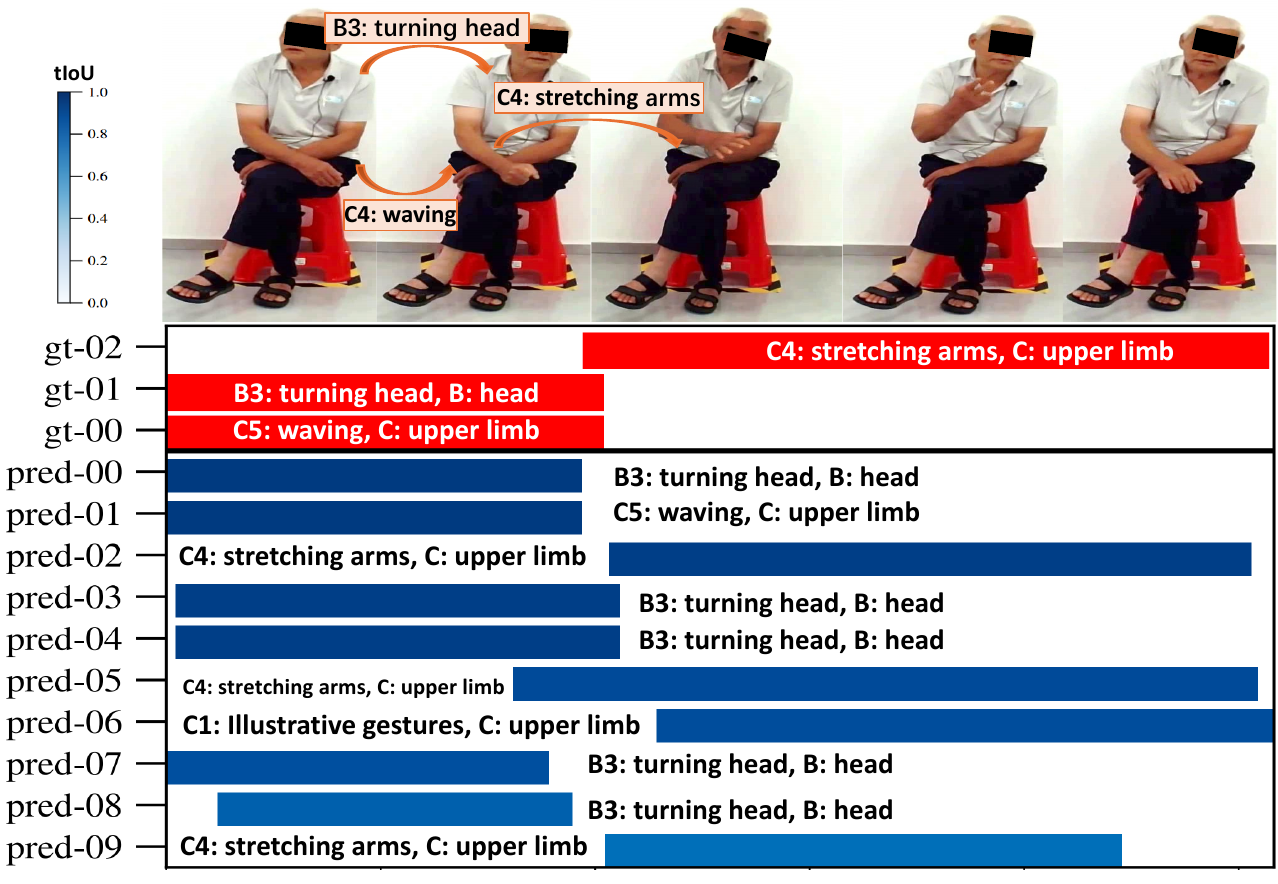}
\caption{\textbf{Qualitative visualization of the prediction.} 
In the vertical axis, gt-$i$ represents the $i$-th ground truth, while pred-$i$ denotes the $i$-th micro-action proposal. 
The color bar represents the tIoU value between the proposal and the ground truth.  
}
\label{fig:vis_results}
\end{figure}

\subsection{Visualization of Prediction}
Additionally, we present the qualitative visualization of the detection results in Fig.~\ref{fig:vis_results}. 
For the first sample, the action of ``D3: shaking legs'' only shows minor visual changes between frames and lasts almost the entire video. Meanwhile, ``B2: shaking head'' sometimes co-occurs. 
From the top 10 predicted proposals, we can see that the proposed method can identify the action boundaries and categories accurately. 
The second example involves the actions of ``C4: stretching arms'', ``C5: waving'' and ``B3: turning head'' across the ``B: head'' and ``C: upper limb'', our method can also detect these co-occurring actions accurately.

\section{Conclusions}
In this paper, we introduced the \textit{\textbf{M}ulti-label \textbf{M}icro-\textbf{A}ction \textbf{D}etection} (\textbf{\textit{MMAD}}) to tackle the challenge of identifying co-occurring micro-actions in real scenarios. 
To facilitate this, we developed the \textit{\textbf{M}ulti-label \textbf{M}icro-\textbf{A}ction-\textbf{52}} (\textbf{\textit{MMA-52}}) dataset, tailored for in-depth analysis and exploration of complex human micro-actions. 
We evaluated 10 baseline models for conventional action detection on the MMA-52 dataset. Besides, we proposed an initial solution with a dual-path spatio-temporal adapter to model the spatial variations and temporal correlations separately. 
The error analysis suggests that there is still a big challenge in detecting micro-actions with low coverage or short length. 
We hope these efforts could encourage the research community to pay more attention to the task of multi-label micro-action recognition and facilitate new advances in human body behavior analysis. 

\section*{Acknowledgments}
This work is supported by National Key R\&D Program of China (NO.2024YFB3311602), Natural Science Foundation of China (62272144), the Anhui Provincial Natural Science Foundation (2408085J040), Anhui Provincial Key Research and Development Project (202304a05020068), the Major Project of Anhui Provincial Science and Technology Breakthrough Program (202423k09020001), the Fundamental Research Funds for the Central Universities (JZ2024HGTG0309, JZ2024AHST0337), and the Earth System Big Data Platform of the School of Earth Sciences, Zhejiang University.

{
\small
\bibliographystyle{ieeenat_fullname}
\bibliography{main}
}

\clearpage

\input{Supplement}

\end{document}

%% file: preamble.tex
\usepackage[dvipsnames]{xcolor}


%% file: math_commands.tex
\usepackage{amsmath,amsfonts,bm}

\def\eqref#1{equation~\ref{#1}}

\def\1{\bm{1}}

\def\mW{{\bm{W}}}
\def\mX{{\bm{X}}}

\DeclareMathAlphabet{\mathsfit}{\encodingdefault}{\sfdefault}{m}{sl}
\SetMathAlphabet{\mathsfit}{bold}{\encodingdefault}{\sfdefault}{bx}{n}

\def\gF{{\mathcal{F}}}

\newcommand{\R}{\mathbb{R}}

\newcommand{\sigmoid}{\sigma}



%% file: tables/main_results.tex
\begin{table*}[t!]
\caption{\textbf{The experimental results on the MMA-52 dataset.} The results are measured by Detection-mAP (\%) at different tIoU thresholds. 
The first block represents the multi-label action detection methods. The second block contains the feature-based methods for TAD, while the third block is end-to-end training methods. 
The best results are marked in \textbf{bold}.
}
\vspace{-0.5em}
\centering
\setlength\tabcolsep{6pt}
\renewcommand\arraystretch{1.1}
\resizebox{0.98\textwidth}{!}{
\begin{tabular}{>{\raggedleft\arraybackslash}p{2.8cm}>{\raggedright\arraybackslash}p{0.8cm}||c||c c c c |c c c c||c}
\hline
\thickhline
\rowcolor{mygray}
& & & \multicolumn{4}{c|}{\texttt{Action-level}} & \multicolumn{4}{c||}{\texttt{Body-level}} &  \\
\rowcolor{mygray}
\multirow{-2}{*}{Method} & & \multirow{-2}{*}{Backbone} & @0.2 & @0.5 & @0.7 & Avg & @0.2 & @0.5 & @0.7 & Avg & \multirow{-2}{*}{\texttt{AVG}}\\
\hline
MS-TCT~\cite{dai2022ms}\!\!\!&\!\!\!\pub{CVPR2021} & I3D & 5.72 & 3.91 & 2.16 & 3.51 & 12.28 & 8.72 & 4.50 & 7.76 & 5.64 \\ 
PointTAD~\cite{tan2022pointtad}\!\!\!&\!\!\!\pub{NeurIPS2021} & I3D & 9.46 & 3.79 & 1.02 & 4.51 & 24.35 & 11.06 & 3.35 & 12.12 & 8.32 \\ \hline
ActionFormer~\cite{zhang2022actionformer}\!\!\!&\!\!\!\pub{ECCV2022} & VideoMAEv2-g & 23.81 &  16.87  &  8.50 &  15.30 & 40.51 & 24.44 &12.20 & 23.99 & 19.65 \\ 
TemporalMaxer~\cite{tang2023temporalmaxer}\!\!\!&\!\!\!\pub{arXiv2023} & VideoMAEv2-g & 25.61 & 17.09 & 7.04 & 15.17 & 43.94 & 26.48 & 11.98 & 25.51 & 20.34 \\
TriDet~\cite{shi2023tridet}\!\!\!&\!\!\!\pub{CVPR2023} & VideoMAEv2-g & 22.41 & 12.06 & 4.59 & 12.45& 35.60 & 19.99 & 7.84 & 19.62  & 16.04 \\
DyFadet~\cite{le2024dyfadet}\!\!\!&\!\!\!\pub{ECCV2024} & VideoMAEv2-g  & 22.17  & 15.19  & 7.96  & 14.17 & 42.52 & 23.34 & 12.55 & 24.93 & 19.55 \\
VideoMamba~\cite{xu2022emotion}\!\!\!&\!\!\!\pub{ECCV2024} & VideoMAEv2-g & 25.34 & 17.55 & 6.90 & 15.21 & 43.43 & 25.45 & 11.17 & 24.08 & 20.01 \\
\hline
TadTR~\cite{liu2022end}\!\!\!&\!\!\!\pub{TIP2022} & SlowFast-R50 & 16.33  & 9.95  & 6.15 & 8.29 & 32.24 & 18.09 & 8.45 & 18.53 & 13.41 \\
Re$^2$TAL~\cite{zhao2023re2tal}\!\!\!&\!\!\!\pub{CVPR2023} & Swin-Tiny & 15.36  & 6.67  & 3.69  & 8.10 & 33.54 & 12.78 & 4.96 & 15.98 & 12.04\\
Re$^2$TAL~\cite{zhao2023re2tal}\!\!\!&\!\!\!\pub{CVPR2023} & SlowFast-101 & 16.15  & 7.10  & 2.93  & 8.10 & 32.39  & 12.15 & 4.57 & 15.38 & 11.74   \\
AdaTAD~\cite{liu2024end}\!\!\!&\!\!\!\pub{CVPR2024} & VideoMAE-S &  24.94  & 16.78 & 10.93  & 16.25 & 45.51 & 27.90 & 7.52 & 27.35 & 21.80  \\  
AdaTAD~\cite{liu2024end}\!\!\!&\!\!\!\pub{CVPR2024} & VideoMAE-B &  28.73 & 19.23 & 8.78 & 17.44 & 49.05 & 28.86 & 7.84 & 28.71 & 23.08\\  
\hline
\hline
\rowcolor[HTML]{f8f9fa}
\multicolumn{2}{c||}{\textbf{\textsc{DSTA} (Ours)}} & VideoMAE-S &  28.05 & 20.40  & 9.03 & 18.16 & 47.14 & 30.02 & 8.37 & 28.70 & 23.43 \\
\rowcolor[HTML]{f8f9fa}
\multicolumn{2}{c||}{\textbf{\textsc{DSTA} (Ours)}} & VideoMAE-B &  \textbf{31.25} &  \textbf{20.87}  & \textbf{11.51} & \textbf{20.30} & \textbf{48.16} & \textbf{32.40} & \textbf{9.42} & \textbf{30.37} & \textbf{25.34}\\
\hline
\end{tabular}}
\vspace{-0.5em}
\label{tab:main_results}
\end{table*}

%% file: Supplement.tex
\appendix
\section*{\centering Overview}
This supplementary material provides more details of the MMA-52 dataset, methodology implementation, ablation studies, and visualization results. 
These topics are organized as follows.

\begin{itemize}

\item \S\ref{sec:app_dataset}: Dataset Details.

\item \S\ref{sec:app_details_method}: Implementation Details.

\item \S\ref{sec:app_exp_abl}: More Experiments.

\item \S\ref{sec:app_exp_vis_pred}: More Visualization Results.

\end{itemize}

\section{Dataset Details}\label{sec:app_dataset}
 
The \textbf{M}ulti-label \textbf{M}icro-\textbf{A}ction-52 (\textbf{MMA-52}) dataset is designed for multi-label micro-action detection. 
Each action instance contains both \texttt{Body-level} and \texttt{Action-level} categories. 
\texttt{Body-level} category denotes the body part of micro-action occurring, including [ ``A: Body'', ``B: Head'', ``C: Upper limb'', ``D: Lower limb'', ``E: Body-hand'', ``F: Head-hand'',  and ``G: Leg-hand''.] 
\texttt{Action-level} category denotes the exact name of micro-actions. 
Taking the body-level category ``A: Body'' as an example, there are 5 action-level categories: ``A1: Shaking body'', ``A2: Turning around'', ``A3: Sitting straightly'', ``A4: Shrugging'' and ``A5: Rising up''. 
Body-level labels and action-level labels are naturally hierarchical structures. 
In summary, there are \textbf{7} \texttt{Body-level} categories and \textbf{52} \texttt{Action-level} categories. 
The detailed descriptions and label ID of each micro-action are the same as the MA-52 dataset~\cite{guo2024benchmarking}.

\section{Implementation details}\label{sec:app_details_method}

\begin{figure}[t]
\centering
\includegraphics[width=1.0\linewidth]{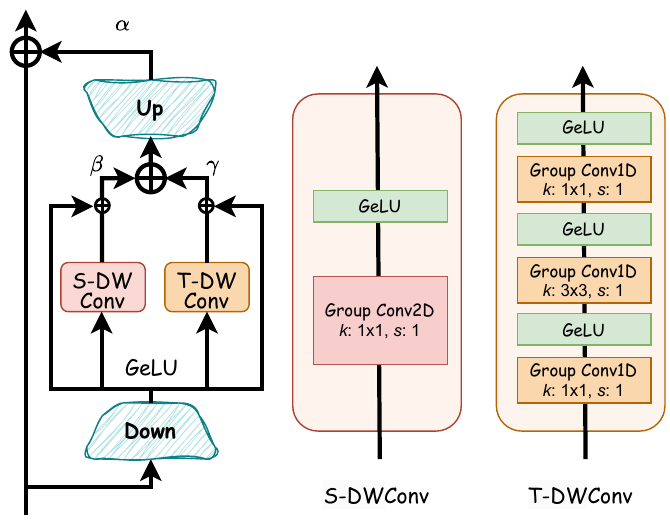}
\caption{\textbf{Overview of the proposed Dual-path Spatial-Temporal Adapter.} ``\texttt{S-DWConv}'' denotes the spatial-path while ``\texttt{T-DWConv}'' denotes the temporal-path}
\label{fig:r_archs}
\end{figure}

As shown in Figure~\ref{fig:r_archs}, we give the detailed arches of the proposed \textbf{D}ual-path \textbf{S}patial-\textbf{T}emporal \textbf{A}dapter (\textbf{DSTA}). For the spatial-path, we use the depth-wise spatial convolution with the kernel size of $1\times 1$ to capture the variation of micro-actions between consecutive frames. 
For the temporal-path, inspired by the bottleneck design in~\cite{sanchez2021affective,howard2017mobilenets}, we design the depth-wise temporal convolution with the kernel size of $\{1, 3, 1\}$. 
After that, we use two separate parameters, $\beta$, and $\gamma$, to dynamically fuse the information from the spatial path and temporal path, respectively. 
The loss optimization of the proposed method is the same as the baseline model~\cite{liu2024end}.

\subsection{Evaluation Metric}
Note that the evaluated methods can only predict the \texttt{action-level} categories. To make a fair comparison, we follow the common practice~\cite{guo2024benchmarking,li2025prototypical,gu2025motion} directly mapping the predicted \texttt{action-level} category to \texttt{body-level} category. 
Then, we use the mAP~\cite{liu2024end} as the evaluation metric to evaluate the performance of the model on the \texttt{action-level} category and \texttt{body-level} category. 
To evaluate the performance at the action level and the body level simultaneously, we report the average value (\texttt{AVG}) derived from the mean of the body-level and action-level scores.

\section{More Experiments}\label{sec:app_exp_abl}

\begin{table}[t]
\centering
\setlength\tabcolsep{8pt}
\caption{\textbf{Ablation study of the position of the adapter.} }
\resizebox{1.0\linewidth}{!}{
\begin{tabular}{ccc|cccc}
\hline\thickhline
\rowcolor{mygray}
\multicolumn{3}{c|}{Layers} & \multicolumn{4}{c}{Action-level Detection-mAP} \\\hline
1-4 & 5-8 & 9-12 & @0.2 & @0.5 & @0.7 & Avg. \\ \hline
\ding{51} & & & 16.71 & 10.38 & 5.31 & 10.38 \\
 & \ding{51} & & 22.07 & 14.11 & 7.51 & 13.83 \\
 &  & \ding{51} & 21.36 & 13.54 & 7.47 & 13.52 \\
\ding{51} & \ding{51} & \ding{51} & \textbf{28.05} & \textbf{20.40} & \textbf{9.03} & \textbf{18.16} \\ \hline
\end{tabular}}
\label{tab:abl_layers}
\end{table}

\subsection{More Ablation Studies}
\noindent \textbf{Ablation studies on the position of the adapter.} 
Here, we evaluate the effect of the position of the adapter. As shown in Table~\ref{tab:abl_layers}, we divided the 12 layers of VideoMAE-S into three stages, namely early (1-4), middle (5-8), and latter (9-12). The results show that the adapter on the middle and latter layers produces better results than the early layers. 
The adapter on all layers achieves the best result of 18.16 in average mAP.

\begin{figure}[t!]
\centering
\begin{subfigure}{\linewidth}
\centering
\includegraphics[width=1.0\linewidth]{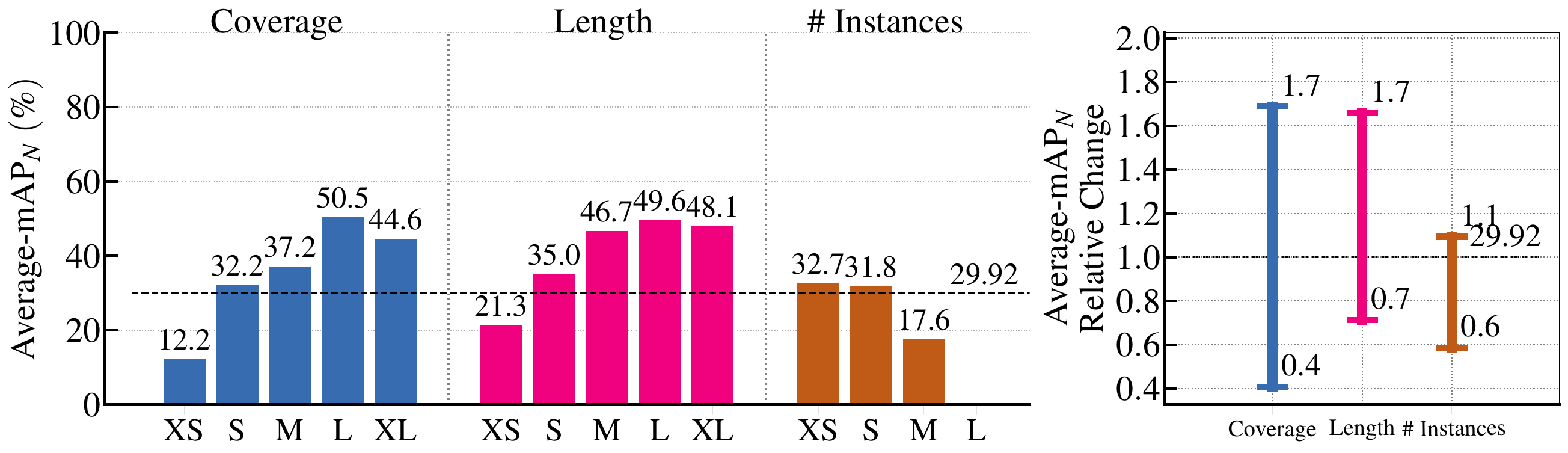}
\caption{\texttt{Action-level} Results}
\end{subfigure}%
\vspace{0.1cm}   
\begin{subfigure}{\linewidth}
\centering
\includegraphics[width=1.0\linewidth]{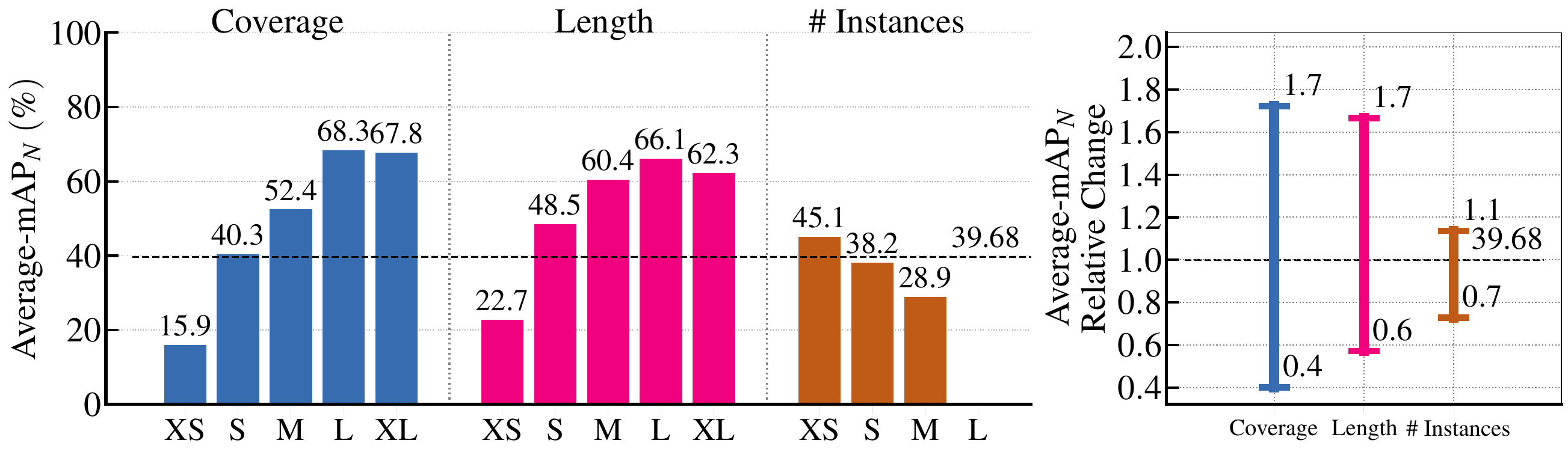}
\caption{\texttt{Body-level} Results}
\end{subfigure}
\caption{\textbf{Sensitivity Analysis.} The left is the normalized mAP at tIoU=0.5, while the right
is the relative normalized mAP change at tIoU=0.5.}
\label{fig:sa}
\end{figure}

\begin{figure}[t!]
\centering
\includegraphics[width=1.0\linewidth]{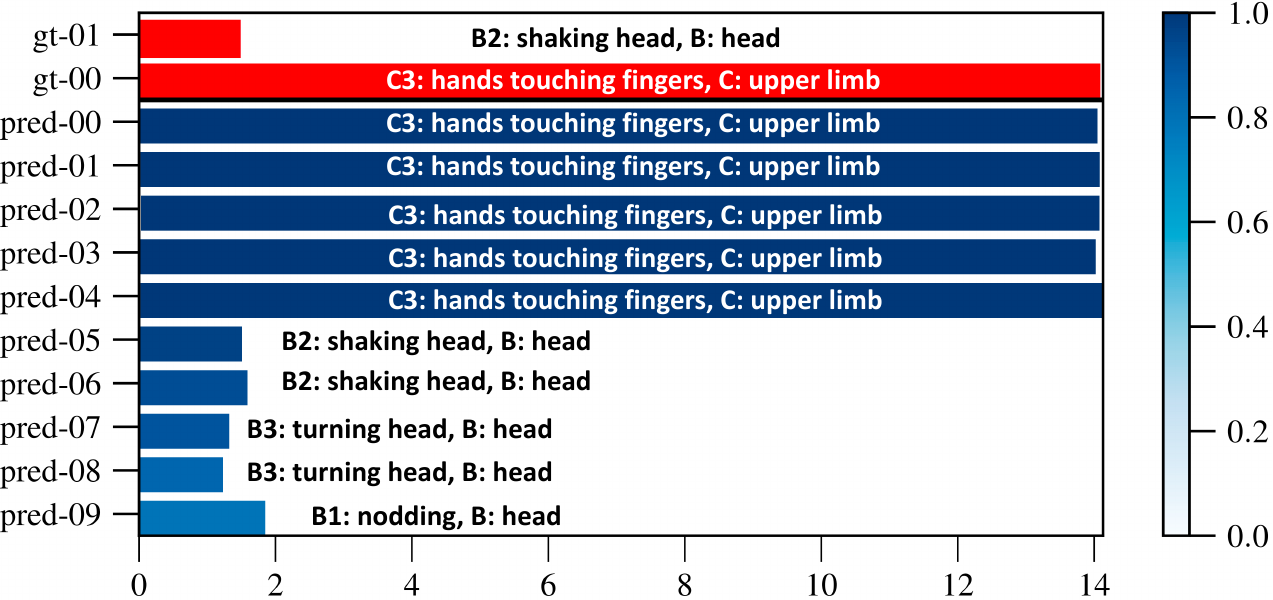}
\includegraphics[width=1.0\linewidth]{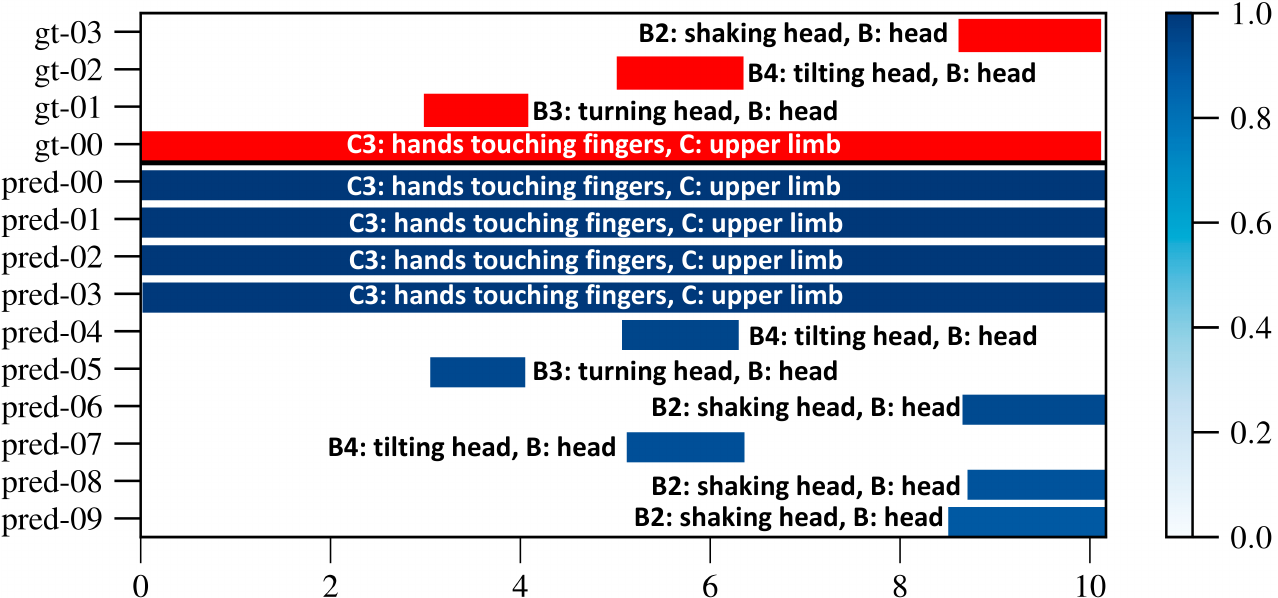}
\includegraphics[width=1.0\linewidth]{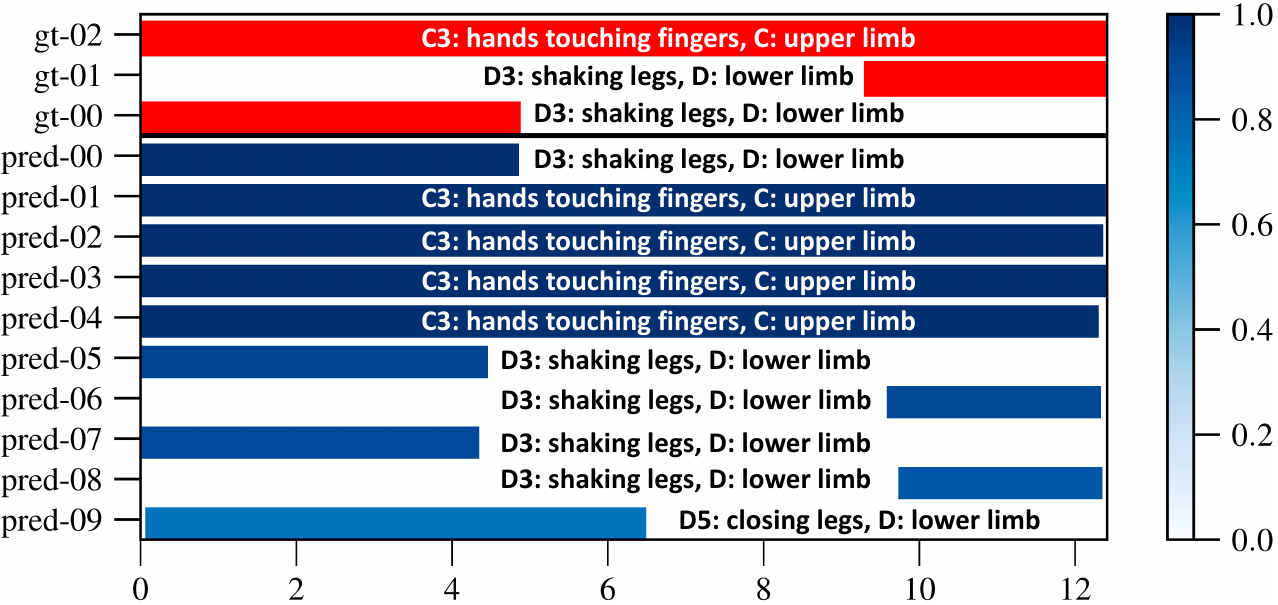}
\caption{\textbf{Top-10 proposals of the proposed method.} In the vertical axis, gt-$i$ represents the $i$-th ground truth, while pred-$i$ denotes the $i$-th micro-action proposal. 
The color bar denotes the tIoU between the proposal and ground truth actions.}
\label{fig:app_visual_pred}
\end{figure}

\subsection{Sensitive Analysis}
As shown in Figure~\ref{fig:sa}, we perform the sensitivity analysis of the proposed model. 
The Average-mAP results are categorized into different buckets by three characteristics, \ie, ``Coverage'', ``Length'', and ``\#Instances''.  ``Coverage'' denotes the ratio of instances within the video,  ``Length'' represents the duration (seconds) of instances, and ``\#Instances'' is the number of instances. 
Following the False Negative analysis settings in the main paper, the range of these characteristics are as follows, \ie, ``Coverage'' refers to [0.0, 0.2, 0.4, 0.6, 0.8, 1.0], ``Length'' refers to [0, 3, 5, 8, 9, INF], and ``\#Instances'' refers to [-1, 2, 5, 7, INF]. 
In each subplot, the left column represents the mAP$_{N}$ and average mAP$_{N}$ for each specific bucket on the MMAD dataset when tIoU=0.5, while the right graph summarizes the left graph by displaying the sensitivity range, calculated as the difference between the maximum and minimum mAP$_{N}$ values. 

From the left column, we can find that: 1) In the characteristics of coverage and length, the average mAP$_{N}$ gradually increases. The lowest performance is the XS bucket and the highest is the L bucket. These results suggest that instances with more coverage or longer length are generally better recognized; 2) In the characteristics of ``\#Instances'', the average mAP$_{N}$ gradually decreases with the increase of action instances. The worst result is in the L bucket. These results indicate that dense instances will lead to lower performance. 
3) The average mAP$_{N}$ at the body-level is better than the action-level, which suggests that the detection of action-level labels is easier. 
From the right column, we have drawn that ``Coverage'' and ``Length'' are the most sensitive factors, with a relative change of 1.7, while ``\#Instances'' is the least sensitive factor, with a relative change of 1.1.
These results indicate the number of instances has a small influence on performance. 

In summary, future work should focus on improving the accuracy of recognizing the instances with shorter coverage and length, as these are the most sensitive factors. In addition, the Average-mAP scores at the body-level are generally higher than the action-level, implying that the body-level information can be used to enhance the detection accuracy at the action-level.

\section{More Visualization Results}\label{sec:app_exp_vis_pred}

As shown in Figure~\ref{fig:app_visual_pred}, we give more prediction results on the proposed MMA-52 dataset. In each subfigure, the red bars denote the ground truth instances, and the blue bars below them are the top 10 predicted proposals. The right color bar is the tIoU value between the proposal and the ground truth. The darker the color, the higher the value of tIoU. These results demonstrate that the proposed method achieves high precision in detecting micro-actions, even those with high co-occurrence and short durations.